\documentclass[runningheads]{llncs}

\usepackage{defs}

\begin{document}
	
\title{Integrating Domain Knowledge: Using Hierarchies to Improve Deep Classifiers}
\titlerunning{Using Hierarchies to Improve Deep Classifiers}
\author{Clemens-Alexander Brust\inst{1}\orcidID{0000-0001-5419-1998} \and
	Joachim Denzler\inst{1,2}}
\institute{Computer Vision Group, Friedrich Schiller University Jena, Jena, Germany \and
	Michael Stifel Center Jena, Jena, Germany}
\maketitle              %
\begin{abstract}
	One of the most prominent problems in machine learning in the age of deep learning is the availability of sufficiently large annotated datasets. For specific domains, \eg animal species, a long-tail distribution means that some classes are observed and annotated insufficiently. Additional labels can be prohibitively expensive, \eg because domain experts need to be involved. However, there is more information available that is to the best of our knowledge not exploited accordingly. 

In this paper, we propose to make use of preexisting class hierarchies like WordNet to integrate additional domain knowledge into classification. We encode the properties of such a class hierarchy into a probabilistic model. From there, we derive a novel label encoding and a corresponding loss function. On the ImageNet and NABirds datasets our method offers a relative improvement of $10.4\%$ and $9.6\%$ in accuracy over the baseline respectively. After less than a third of training time, it is already able to match the baseline's fine-grained recognition performance. Both results show that our suggested method is efficient and effective.
	
\end{abstract}

\section{Introduction}
\begin{figure}[t]
  \begin{center}
  \subfloat[Class Set]{%
    \includegraphics[scale=0.3]{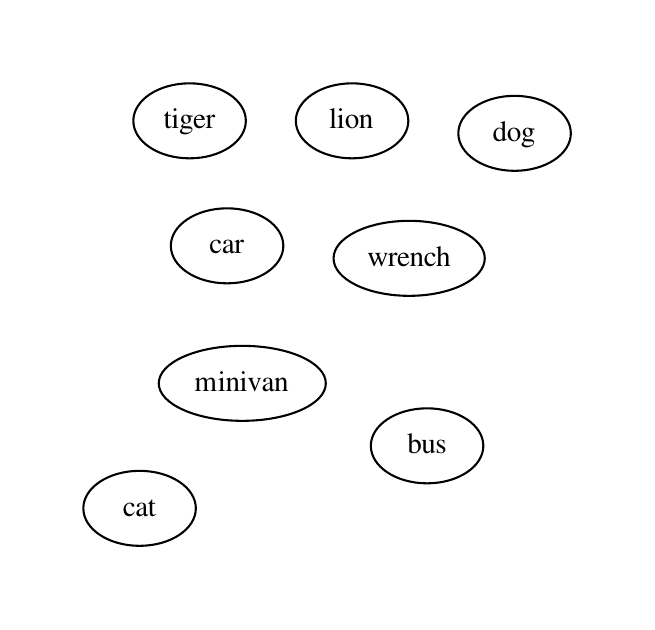}%
    }%
  \subfloat[Class Hierarchy]{%
    \includegraphics[scale=0.3]{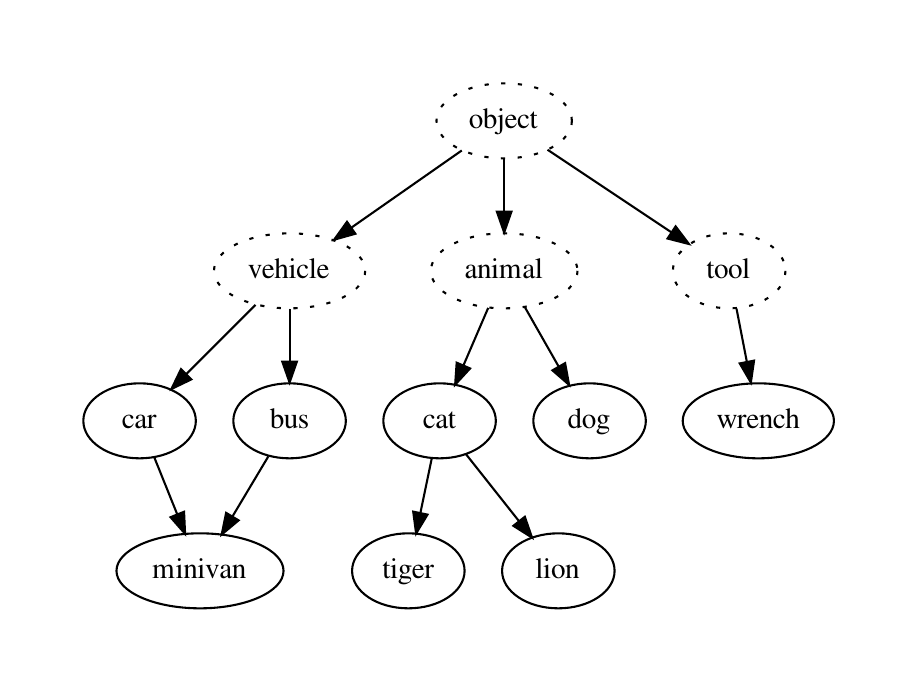}%
    }%
  \caption{Comparison between a loose set of independent classes and a class hierarchy detailing inter-class relations.\vspace{-15pt}}
  \label{fig:teaser}
  \end{center}
\end{figure}

In recent years, \glspl{cnn} have achieved outstanding performance
in a variety of machine learning tasks, especially in computer vision, such as
image classification \cite{He2015Res,Krizhevsky2012CNN} and semantic segmentation \cite{Long2014FCN}.
Training a \gls{cnn} from scratch in an end-to-end fashion not only requires considerable
computational resources and experience, but also large amounts of labeled training data \cite{Sun2017Data}.
Using pre-trained \gls{cnn} features \cite{Razavian2014Features}, adapting
existing \glspl{cnn} to new tasks \cite{Hoffman2014LSDA} or performing data augmentation
can reduce the need for
labeled training data, but may not always be applicable or effective.

For specific problem domains, \eg with a long-tailed distribution of samples over classes, the amount of labeled training
data available is not always sufficient for training a \gls{cnn} to reasonable performance.
When unlabeled data already exists, which is not always the case,
active learning \cite{Settles2009ALS}
to select valuable instances for labeling may be applied.
However, labels still have to be procured which is not always feasible.

Besides information from more training data, domain knowledge in the form of high-level information about the structure
of the problem can be considered. In contrast to annotations of training data, this kind of domain knowledge is already available
in many cases from projects like
iNaturalist \cite{VanHorn2017iNat}, Visual Genome \cite{Krishna2017VG}, Wikidata \cite{Vrandevcic2014Wikidata}
and WikiSpecies\footnote{\url{https://species.wikimedia.org/wiki/Main_Page}}.

In this paper, we use class hierarchies, \eg WordNet \cite{Fellbaum1998WordNet}, as
an example of domain knowledge. In contrast to approaches based on attributes, where annotations
are often expected to be per-image, class hierarchies offer the option of domain knowledge integration on the highest level with the least additional annotation effort.
We encode the properties of such a class hierarchy into a probabilistic model that is based on common assumptions around hierarchies. From there, we derive a special label encoding together with a corresponding loss function.
These components are applied to a \gls{cnn} and evaluated systematically.
Our main \textbf{contributions} are: (i) a deep learning method based on a probabilistic model to improve existing classifiers by adding a
class hierarchy which (ii) works with any form of hierarchy representable using a
\glsfirst{dag}, \ie does not require a tree hierarchy. We evaluate our method in
experiments on the CIFAR-100 \cite{Krizhevsky2009CIFAR}, ImageNet and NABirds \cite{VanHorn2015NAB} datasets to represent problem
domains of various scales.

\section{Related Work}
Hierarchical methods have been subject of extensive
research in image categorization. A given class hierarchy can be used
explicitly to build a hierarchical classifier
\autocite{hwang_learning_2011,marszalek_semantic_2007},
to regularize a preexisting model
\autocite{fergus_semantic_2010,srivastava_discriminative_2013},
to construct an embedding space
\autocite{barz_hierarchy-based_2019,faghri_vse++:_2017,frome_devise:_2013,
	hwang_unified_2014},
in metric learning-based methods
\autocite{hwang_learning_2011,verma_learning_2012,zhang_embedding_2016}
or, to construct a probabilistic model
\autocite{deng_large-scale_2014, gaussier_hierarchical_2002}.

Leveraging external semantic information for performance improvements,
also called knowledge transfer, has
been studied in the context of text categorization
\autocite{benkhalifa_integrating_2001} as well as
visual recognition
\autocite{hwang_discriminative_2013,rodner_one-shot_2010,wu_image_2018}.
Attributes are also considered as a knowledge source in
\autocite{hwang_unified_2014}.
While improvements are generally expected when using such methods,
disagreements between visual and semantic similarity may introduce
new errors \autocite{brust_not_2018}. Alternatively, visual hierarchies
can be learned
\autocite{bart_unsupervised_2008,yan_hd-cnn:_2015} or used implicitly
\autocite{bilal_convolutional_2018,deng_what_2010}.
An extreme case of knowledge transfer is zero-shot learning, where some categories have zero
training examples
\autocite{huo_zero-shot_2018,rohrbach_transfer_2013}.

Our work is most closely related to \autocite{marszalek_semantic_2007}
in that we consider similar individual classification problems. However,
instead of
their step-by-step approach using binary classifiers,
our probabilistic model is evaluated globally for inference. A similar
approach is also used in \autocite{deng_large-scale_2014}, where a relations
between classes such as subsumption and mutual exclusion are extracted from
a hierarchy and then used to condition a graphical model.

\paragraph{Hierarchical Data}
Typical image classification datasets rarely offer hierarchical information. There are exceptions
such as the iNaturalist challenge dataset \cite{VanHorn2017iNat} where a class hierarchy is
derived from biological taxonomy. Exceptions also include specific hierarchical classification
benchmarks, \eg \cite{Partalas2015LSHTC,Torralba2008Tiny} as well as datasets where the labels
originate from a hierarchy such as ImageNet \cite{Deng2009ImageNet}.
The Visual Genome dataset \cite{Krishna2017VG} is another notable exception, with available metadata
including attributes, relationships, visual question answers, bounding boxes and more, all
mapped to elements from WordNet.

To augment existing non-hierarchical datasets, class hierarchies can be used.
For a typical object classification scenario, concepts from the WordNet database
\cite{Fellbaum1998WordNet} can be mapped to object classes. WordNet contains
nouns, verbs and adjectives that are grouped into \emph{synsets} of synonymous concepts.
Relations such as hyponymy (\textbf{is-a}), antonymy (\textbf{is-not}), troponymy (\textbf{is-a-way-of}) and meronymy (\textbf{is-part-of}) are encoded in a graph structure
where synsets are represented by nodes and relations by edges respectively. In this paper,
we use the hyponymy relation to infer a class hierarchy.

\section{Method}
In this section, we propose a method to adapt existing classifiers
to hierarchical classification. We start by acquiring a hierarchy and then define a probabilistic
model based on it. From this probabilistic model, we derive an encoding and a loss function
that can be used in a machine learning environment.

\subsection{Class Hierarchy}
For our model, we assume that a hierarchy of object categories is supplied,
\eg from a database such as WordNet \cite{Fellbaum1998WordNet} or WikiSpecies.
It is modeled in the form of a graph $W=(S,h)$, where $S$ denotes the set of all possible
object categories, called \emph{synsets} in the WordNet terminology.
These are the nodes of the graph. Note that $S$ is typically a superset of the dataset
categories $C \subseteq S$, since parent categories are included to connect existing categories,
\eg \texttt{vehicle} is a parent of \texttt{car} and \texttt{bus}, but not originally part
of the dataset.

A hyponymy relation $h \in S \times S$ over the classes, which can be
interpreted as directed edges in the graph, is also given. For example, $(s,s')\in h$ means that
$s'$ is a hyperonym of $s$, or $s$ is a hyponym of $s'$, meaning $s$ \textbf{is-a} $s'$.
In general, the \textbf{is-a} relation is transitive. However, WordNet only models direct
relationships between classes to keep the graph manageable and to represent different
levels of abstraction as graph distances. The relation is also irreflexive and
asymmetric. 

For the following section, we assume that $W$ is a \glsfirst{dag}. However,
the WordNet graph is commonly reduced to a tree, for example by using a voting algorithm \cite{Torralba2008Tiny}
or selecting task-specific subsets that are trees \cite{Deng2009ImageNet}. In this paper, we work on the \gls{dag} directly.

\subsection{Probabilistic Model}
Elements of a class hierarchy are not always mutually exclusive, \eg a \texttt{corgi} is also a \texttt{dog} and an \texttt{animal} at the same time.
Hence, we do not model the class label as one categorical random variable, but
assume multiple independent Bernoulli variables $Y_{s}, s \in S$ instead. 
Formally, we model the probability of any class $s$ occurring independently (and thus allowing even multi-label
scenarios), given an example $x$:
\begin{equation}
P(Y_{s}=1 | X=x),
\end{equation}
or, more concisely,
\begin{equation}
P(Y_s^+ | X).
\label{eqn:themodel}
\end{equation}

The aforementioned model on its own is overly flexible considering the problem at hand, since
it allows for any combination of categories co-occurring. At this point, assumptions are similar
to those behind a one-hot encoding.
However, from the common definition of a hierarchy, we can infer a few additional properties to restrict the model.

\paragraph{Hierarchical decomposition}
A class $s$ can have many independent parents $S'=s'_1,\ldots,s'_n$. We choose $Y_{S'}^+$ to denote
an observation of at least one parent and $Y_{S'}^-$ to indicate that no parent class
has been observed:
\begin{eqnarray*}
Y_{S'}^+ & \Leftrightarrow Y_{s'_1}^+ \vee \ldots \vee Y_{s'_n}^+   \Leftrightarrow Y_{s'_1}=1 \vee \ldots \vee Y_{s'_n}=1,\\
Y_{S'}^- & \Leftrightarrow Y_{s'_1}^- \land \ldots \land Y_{s'_n}^- \Leftrightarrow Y_{s'_1}=0 \land \ldots \land Y_{s'_n}=0.
\end{eqnarray*}
Based on observations $Y_{S'}$, we can decompose the model from \cref{eqn:themodel}
in a way to capture the hierarchical nature. We start by assuming a marginalization
of the conditional part of the model over the parents $Y_{s'}$:
\begin{equation}
\begin{split}
P(Y_s^+|X) &= P(Y_s^+|X, Y_{S'}^+)P(Y_{S'}^+|X)\\
&+ P(Y_s^+|X, Y_{S'}^-)P(Y_{S'}^-|X).
\end{split}
\label{eqn:decompose}
\end{equation}
The details of this decomposition are given in the supplementary material.

\paragraph{Simplification}
We now constrain the model and add assumptions to better reflect the hierarchical problem.
If none of the parents $S'=s'_1,\ldots,s'_n$ of a class $s$ occur, we assume the
probability of $s$ being observed for any given example to be zero:
\begin{equation}
P(Y_s^+ | X, Y_{S'}^-) = P(Y_s^+ | Y_{S'}^-) = 0.
\label{eqn:hcritzero}
\end{equation}
This leads to a simpler hierarchical model, omitting the second half of \cref{eqn:decompose}
by setting it to zero:
\begin{equation}
P(Y_s^+|X) = P(Y_s^+|X, Y_{S'}^+)P(Y_{S'}^+|X).
\end{equation}

\paragraph{Parental independence}
To make use of recursion in our model, we require the random variables $Y_{s'_1}, \ldots, Y_{s'_n}$
to be independent of each other in a naive fashion. Using the definition of $Y_{S'}^+$, we
derive:
\begin{equation}
P(Y_{S'}^+|X) = 1 - \prod_{i=1}^{|S'|}{1-P(Y_{s'_i}^+|X)}.
\label{eqn:parindep}
\end{equation} 

\paragraph{Parentlessness}
In a non-empty \gls{dag}, we can expect there to be at least one node with no incoming edges, \ie a
class with no parents. In the case of WordNet, there is exactly one node with no parents, the root synset \texttt{entity.n.01}.
A marginalization over parent classes does not apply there. We assume that all observed
classes are children of \texttt{entity} and thus set the probability to one for a class without parents:
\begin{equation}
P(Y_s^+|X,S'=\emptyset) = 1.
\label{eqn:entity}
\end{equation}
Note that this is not reasonable for all hierarchical classification problems. If the
hierarchy is composed of many disjoint components, $P(Y_s^+|X,S'=\emptyset)$ should
be modeled explicitly. Even if there is only a single root, explicit modeling could
be used for tasks such as novelty detection.

\subsection{Inference}
\label{sec:inference}
The following section describes the details of the inference process in our model.

\paragraph{Restricted Model Outputs}
\label{sec:mlnp}
Depending on the setting, when the model is used for inference, the possible outputs can be restricted
to the classes $C$ that can actually occur in the dataset as opposed to all modeled
classes $S$ including parents that exist only in the hierarchy. This assumes a known class set at test time as opposed to an open-set problem.
We denote this setting \emph{\gls{mlnp}} and the unrestricted alternative \emph{\gls{anp}}.

\paragraph{Prediction} To predict a \emph{single} class $s$ given a specific example $x$, we look for the class where
the joint probability of the following observations is high: (i) the class $s$ itself occurring ($Y_s^+$)
and (ii) none of the children $S''=s''_1,\ldots,s''_m$ occurring ($Y_{S''}^-$):

\begin{equation}
    s(x) = \argmax_{s \in C \subseteq S}{P(Y_s^+|X)}P(Y_{S''}^-|X,Y_s^+). %
\end{equation}
Requiring the children to
be pairwise independent similar to \cref{eqn:parindep}, inference is performed in the following way:
\begin{equation}
   s(x) =  \argmax_{s \in C \subseteq S}{P(Y_s^+|X)\prod_{i=1}^{|S''|}{1-P(Y_{s''_i}^+|X,Y_s^+)}}.
\end{equation}
Because $P(Y_s^+|X)$ can be decomposed according to \cref{eqn:decompose} and expressed as a product (cf. \cref{eqn:parindep}),
we infer using:
\begin{equation}
  \begin{split}
   s(x) = \argmax_{s \in C \subseteq S}\,
   P(Y_s^+|X, Y_{S'}^+)%
   &\cdot \underbrace{(1 - \prod_{i=1}^{|S'|}{1-P(Y_{s'_i}^+|X)})}_{\text{Parent nodes }S'}
   \cdot \underbrace{\prod_{i=1}^{|S''|}{1-P(Y_{s''_i}^+|X,Y_s^+)}}_{\text{Child nodes }'S}.
   \label{eqn:argmax}
  \end{split}
\end{equation}
Again, $P(Y_{s'_i}^+|X)$ can be decomposed. This decomposition is performed recursively
following the scheme laid out in \cref{eqn:decompose} until a parentless node is reached  (cf. \cref{eqn:entity}).

\subsection{Training}
\label{sec:loss}
\begin{figure}[t]
  \begin{center}
  \subfloat[Encoding $e(y)$]{%
    \includegraphics[scale=0.3]{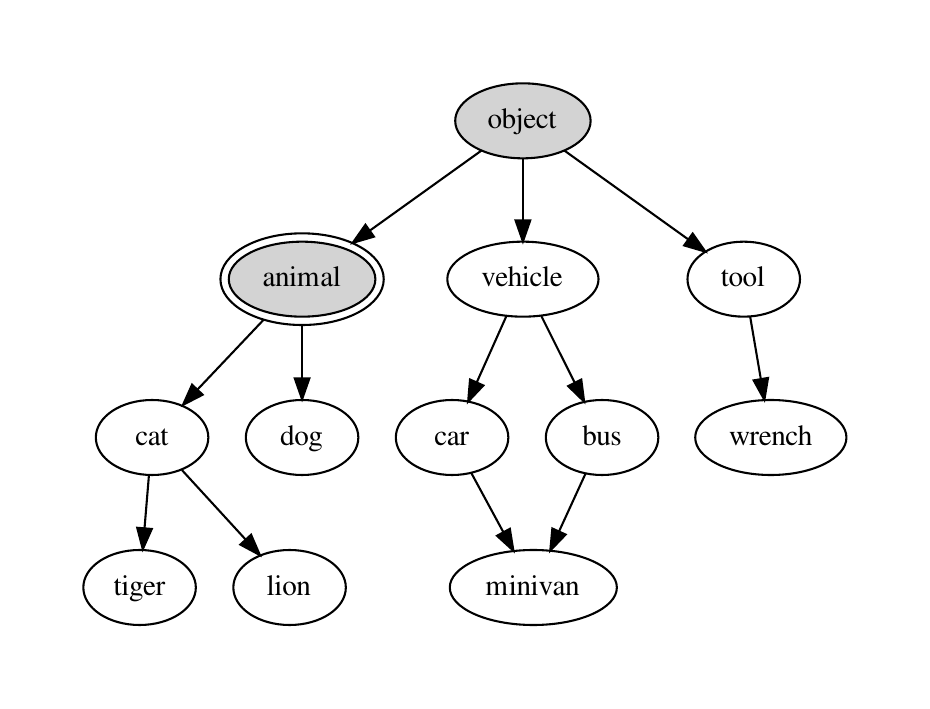}%
    }%
  \subfloat[Loss Mask $m(y)$]{%
    \includegraphics[scale=0.3]{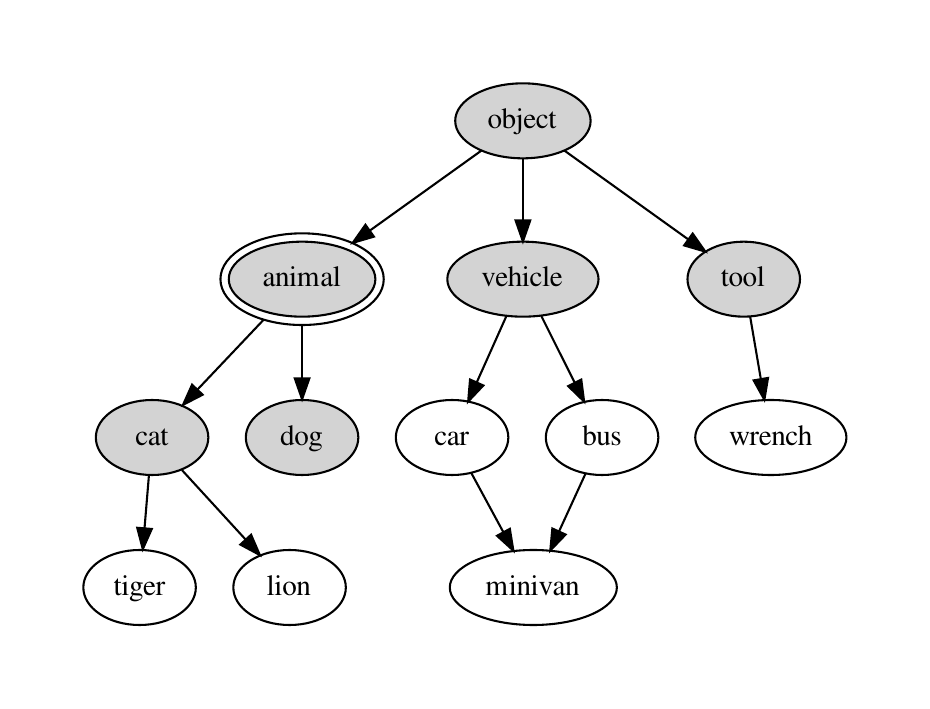}%
    }%
  \caption{Hierarchical encoding and loss mask for $y=\texttt{animal}$. Shaded
  nodes represent $1$ and light nodes $0$ respectively.}
  \vspace{-15pt}
  \label{fig:henc}
  \end{center}
\end{figure}
In this section, we describe how to implement our proposed model in a machine learning context.
Instead of modeling the probabilities $P(Y_s^+|X)$ for each class $s$ directly, we want to estimate
the conditional probabilities $P(Y_s^+|X, Y_{S'}^+)$. This changes each individual estimator's task
slightly, because it only needs to discriminate among siblings and not all classes. It also enables the implementation of
the hierarchical recursive inference used in \cref{eqn:argmax}.

The main components comprise of a label encoding $e: S \to \{0,1\}^{|S|}$ as well as a special loss function.
A label $y \in S$ is encoded using the hyponymy relation $h \in S \times S$, specifically its transitive
closure $\mathcal{T}(h)$, and the following function:
\begin{equation}
e(y)_s = \begin{cases}
1 & \text{if }y = s\text{ or }(y,s) \in \mathcal{T}(h),\\
0 & \text{otherwise}. 
\end{cases}
\end{equation}

A machine learning method can now be used to estimate encoded labels directly. However,
a suitable loss function needs to be provided such that the conditional nature of each
individual estimator is preserved. This means that, given a label $y$, a component $s$ should be trained only
if one of its parents $s'$ is related to the label $y$ by $\mathcal{T}(h)$, or if $y$ is one of its parents.
We encode this requirement using a \emph{loss mask} $m: S \to \{0,1\}^{|S|}$, defined by the
following equation:

\begin{equation}
m(y)_s = \begin{cases}
1 & \text{$y=s$ or} \\
 & \exists (s,s') \in h: y = s' \text{ or }(y,s') \in \mathcal{T}(h),\\
0 & \text{otherwise}. 
\end{cases}
\end{equation}

\Cref{fig:henc} visualizes the encoding $e(y)$ and the corresponding loss mask $m(y)$ for a
small example hierarchy. Using the encoding and loss mask, the
complete loss function $\mathcal{L}$ for a given data point $(x,y)$ and estimator $f: X \to \{0,1\}^{|S|}$ is then defined
by the the masked mean squared error (alternatively, a binary cross-entry loss can be used):
\begin{equation}
\mathcal{L}_f(x,y) = m(y)^T(e(y) - f(x))^2.
\end{equation}

The function $f(x)_s$ is then used to estimate the conditional probabilities $P(Y_s^+|X, Y_{S'}^+)$.
Applying the inference procedure in \cref{sec:inference}, a prediction is made
using the formula in \cref{eqn:argmax} and substituting $f(x)_s$ for $P(Y_s^+|X, Y_{S'}^+)$.

\section{Experiments}
\label{sec:expdata}
We aim to assess the effects of applying our method on three different scales of problems, using the
following datasets:
\paragraph{CIFAR-100}
For our experiments, we want to work with a dataset that does not directly supply hierarchical labels,
but where we can reasonably assume that an underlying hierarchy exists.
The CIFAR-100 dataset \cite{Krizhevsky2009CIFAR} fulfills this requirement. Because there are only 100
classes, each can be mapped to a specific synset in the WordNet hierarchy without
relying on potentially faulty automation. Direct mapping is not always possible, \eg \texttt{aquarium\_fish}, which
doesn't exist in WordNet and was mapped to \texttt{freshwater fish.n.01} by us. This process is a potential error source.

The target hierarchy is composed in three steps. First, the synsets mapped from all CIFAR-100
classes make up the foundation. Then, parents of the synsets are added in a recursive fashion.
With the nodes of the graph complete, directed edges are determined using the WordNet
hyponymy relation.
Mapping all classes to the WordNet synsets results in 99 classes being mapped to leaf nodes
and one class to an inner node (\texttt{whale}). In total, there are 267 nodes as a result of
the recursive adding of hyperonyms. 

\paragraph{ImageNet} 
Manually mapping dataset labels to WordNet synsets is a potentional source of errors. An ideal dataset would
use WordNet as its label space. Because of WordNet's popularity, such
datasets exist, \eg ImageNet \cite{Deng2009ImageNet} and 80 Million Tiny Images \cite{Torralba2008Tiny}.
We use ImageNet, specifically the dataset of the 2012 ImageNet Large Scale Visual Recognition Challenge.
It contains around 1000 training images each for
1000 synsets.
All 1000 synsets are leaf nodes in the resulting hierarchy with a total of 1860 nodes.

\paragraph{NABirds}
Quantifying performance on object recognition datasets such as CIFAR and
ImageNet is important to prove the general usefulness of a method. However, more niche applications
such as fine-grained recognition stand to benefit more from improvements because the availability
of labeled data is much more limited.
We use the NABirds dataset \cite{VanHorn2015NAB} to verify our method in a fine-grained recognition
setting. NABirds is a challenge where 555 categories of North American birds have to be differentiated.
These categories are comprised of 404 species as well as several variants of sex, age and plumage.
It contains 48,562 images split evenly into training and validation sets.
Annotations include not only image labels, but also bounding boxes and parts.
Additionally, a class hierarchy based on taxonomy is supplied. 
It contains 1010 nodes, where all of the 555 visual categories are leaf nodes.

\subsection{Experimental Setup}
\paragraph{Convolutional Neural Networks}
\label{sec:expsetup}
For our experiments on the CIFAR-100 dataset, we use a ResNet-32 \cite{He2015Res}
in the configuration originally designed for CIFAR. The network is initialized randomly as specified
in \cite{He2015Res}.

We use a minibatch size of 128 and the adaptive stochastic optimizer Adam \cite{Kingma2014Adam} with a
constant learning rate of $0.001$ as recommended by the authors. Although SGD can lead to better
performance of the final models, its learning rate is more dependent on the range of the loss function.
We choose an adaptive optimizer to minimize the influence of different ranges of loss values.

In our NABirds and ImageNet experiments, we use a ResNet-50 \cite{He2015Res,He2016Identity}
 because of the larger image size and overall scale of the dataset.
The minibatch size is reduced to 64 and training is extended to $120,000$ steps for NABirds and $234,375$ steps for ImageNet.
We crop all NABirds images using the given bounding box annotations and resize them to $224\times 224$ pixels.

All settings use random shifts of up to 4 pixels for CIFAR-100 and
up to 32 pixels for NABirds and ImageNet as well as random horizontal flips during training. All images
are normalized per channel to zero mean and standard deviation one, using parameters estimated over the
training data. Code will be published along with the paper.
We choose our ResNet-50 and ResNet-32 baselines to be able to judge effects across datasets, which would not be possible
when selecting a state-of-the-art method for each dataset individually. Furthermore, the moderately sized
architecture enables faster training and therefore more experimental runs compared to a high performing
design such as PNASNet \cite{Liu2017PNASNet}.

\paragraph{Evaluation}
We report the overall accuracy, not normalized w.r.t class instance counts.
Each experiment consists of six random initializations per method for the CIFAR-100
dataset and three for the larger-scale NABirds and ImageNet datasets, over which
we report the average.
We choose to compare the methods using a measure that does not
take hierarchy into account to gauge the effects of adding hierarchical data to a task
that is not normally evaluated with a specific hierarchy in mind. Using a hierarchical measure would
achieve the opposite: we would measure the loss sustained by omitting hierarchical data.

\subsection{Overall Improvement --- ImageNet}
\label{sec:exp-in}
\label{sec:exp-first}

\begin{figure}
\centering
\includegraphics{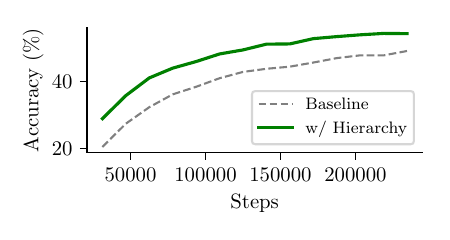}
\vspace{-15pt}
\caption{Accuracy on the ImageNet validation set over time. Our hierarchical training method gains accuracy faster than the
flat classifier baseline. We report overall classification accuracy in percent.}
\vspace{-15pt}
\label{fig:expingraph}
\end{figure}

In this experiment, we quantify the effects of using our hierarchical classification method
to replace the common combination of one-hot encoding and mean squared error loss function.
We use ImageNet, specifically the ILSVRC2012 dataset. This is a classification challenge with 1000 classes
whose labels are
taken directly from the WordNet hierarchy of nouns. 

\Cref{fig:expingraph} shows the evolution over time of accuracy on the validation set.
After around 240,000 gradient steps, training converges. The one-hot baseline reaches
a final accuracy of $49.1\%$, while our method achieves $54.2\%$ with no changes to training
except for our loss function and hierarchical encoding. This is a relative improvement of $10.4\%$.

While an improvement of accuracy at the end of training is always welcome,
the effects of hierarchical classification more drastically show in the change in accuracy over
time. The strongest improvement is observed during the first training steps.
After training for 31250 steps using our method, the network already performs with $28.9\%$ accuracy.
The one-hot baseline matches this performance after 62500 gradient steps, taking twice as long.
The baseline's final accuracy of $49.1\%$ is matched by our method after only $125,000$ training
steps, resulting in an overall training speedup of $1.88\text{x}$.
%
%

%
%

\subsection{Speedup --- CIFAR-100}

\begin{figure}
\centering
\includegraphics{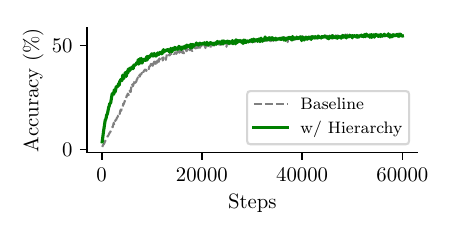}
\vspace{-15pt}
\caption{Results on the CIFAR-100 validation set. Our hierarchical training method gains accuracy faster than the
flat classifier baseline. We report overall classification accuracy in percent.}
\vspace{-15pt}
\label{fig:expcifargraph}
\end{figure}

We report the accuracies on the validation set as they develop during training in
\cref{fig:expcifargraph}. As training converges, we observe almost no difference
between both methods, with our hierarchical method reaching $54.6\%$ and the one-hot
encoding baseline at $55.4\%$. However, the methods differ strongly in the way that accuracy is
achieved. After the first $500$ steps, our hierarchical classifier already predicts $10.7\%$ of
the validation set correctly, compared to the baseline's $2.8\%$. It takes the baseline another
1600 steps to match $10.7\%$, or $4.2$ times as many steps.

This advantage in training speed is very strong during initial training, but becomes smaller
over time. %
After the first half of training, the difference between both methods vanishes almost completely.

\subsection{Fine-Grained Recognition --- NABirds}

%
To evaluate the performance of our hierarchical method in a more specific setting, we
use the NABirds dataset \cite{VanHorn2015NAB}, a fine-grained recognition challenge where
the task is to classify 555 visual categories of birds. A hierarchy is given by the dataset.
We observe results similar to the ImageNet dataset (see \cref{sec:exp-in}), where
our method leads to an improvement in both training speed and overall accuracy.
The one-hot baseline converges to an accuracy of $56.5\%$. Our hierarchical
classifier reaches $61.9\%$ after the full $120,000$ steps of training. It already
matches the baseline's final accuracy at $39,000$ iterations, reducing training
time to less than a third. The relative improvement with full training is $9.6\%$.

\subsection{Overview and Discussion}
\label{sec:exp-overview}
\begin{table}[t]
\centering
\caption{Results Overview.}
\label{tbl:expoverview}
\begin{tabular}{l|l||c|c||c|c}
 &  & \multicolumn{2}{c||}{Accuracy (\%)} & \multicolumn{2}{c}{Speedup w/Hierarchy}\\
Dataset       &  \# of classes  &     Baseline     &       w/Hierarchy        & Overall & Initial \\\hline
CIFAR-100     &  100     & $\textbf{55.4} \pm 0.84$ &  $54.6 \pm 1.03$ & ---            &  $7.00$\\
NABirds       &  555     & $56.5 \pm 0.49$ &  $\textbf{61.9} \pm 0.27$ & $3.08$ &  $10.00$\\
ILSVRC2012  &  1000    & $49.1 \pm 0.33$  & $\textbf{54.2} \pm 0.04$ & $1.88$ & ---
\end{tabular} 
\vspace{-15pt}
\end{table}

\Cref{tbl:expoverview} provides the most important facts for each dataset. We report
the accuracy at the end of training for the one-hot baseline as well as our
method. Overall speedup indicates how much faster in terms of training steps
our hierarchical method achieves the end-of-training accuracy of the baseline.
Initial speedup looks at the accuracy
delivered by our method after the first validation interval. We then measure
how much longer the baseline needs to catch up.

On all 3 datasets, the initial training is faster using our method. However,
we only observe an improvement in classification accuracy on ImageNet
and NABirds. With CIFAR-100, the benefits of adding hierarchical information
are limited to training speed. There are a few possible
explanations for this:

First, the CIFAR-100 dataset is the only dataset that requires a manual mapping to
an external hierarchy, whereas the other datasets either supply one or have
labels directly derived from one. The manual mapping is a possible error source
and as such, could explain the observation, as could the small image size.

The second possible reason lies in the difference between semantic similarity
and visual similarity \cite{brust_not_2018,Deselaers2011Visual}. Semantic similarity relates two classes using their
meaning. It can be extracted from hierarchies such as WordNet \cite{Fellbaum1998WordNet},
for example by looking at distances in the graph. 
Visual similarity on the other hand relates images that look alike, regardless
of the meaning behind them. When classifying, we group images by semantics, even if they share no visual characteristics.
Adding information based on only semantics can thus lead to problems.

\section{Conclusion}
We present a method to modify existing deep classifiers such that knowledge
about relationships between classes can be integrated. The method is derived
from a probabilistic model that is itself based on our understanding of the
meaning of hierarchies. Overall, it is just one example of the integration
of domain knowledge in an otherwise general method. One could also consider
our method a special case of learning using privileged information
\cite{Vapnik2009LUPI}.

Our method can improve classifiers by utilizing information that is freely
available in many cases such as WordNet \cite{Fellbaum1998WordNet} or
WikiSpecies.
There are also datasets which include a hierarchy that is ready to use \cite{Deng2009ImageNet,VanHorn2015NAB}.

Further research should focus on the data insufficiency aspect and quantify the data reduction
made possible by our method on small datasets, and compare the sample
efficiency to the baseline for artificially reduced datasets as well as
alternatives such as data augmentation.

\printbibliography
\end{document}